\documentclass[conference]{IEEEtran}
\IEEEoverridecommandlockouts
\usepackage{hyperref}
\hypersetup{colorlinks,linkcolor={green},citecolor={green},urlcolor={red}}  
\usepackage{lineno,hyperref}

 \usepackage{booktabs,tabulary,lipsum}
 \usepackage{dcolumn,tipa}
 \newcolumntype{d}{D{.}{.}{6.5}}
 \usepackage{siunitx}
 \usepackage[usestackEOL]{stackengine}
 \usepackage{multirow}

\usepackage{cite}
\usepackage{amsmath,amssymb,amsfonts}
\usepackage{algorithmic}
\usepackage{graphicx}
\usepackage{textcomp}
\usepackage{bbm}
\usepackage{xcolor}
\def\BibTeX{{\rm B\kern-.05em{\sc i\kern-.025em b}\kern-.08em
    T\kern-.1667em\lower.7ex\hbox{E}\kern-.125emX}}
\begin{document}

\title{TransPose: A Transformer-based 6D Object Pose Estimation Network with Depth Refinement\\

\thanks{Identify applicable funding agency here. If none, delete this.}
}

\author{
\IEEEauthorblockN{Mahmoud Abdulsalam \IEEEauthorrefmark{5}, and Nabil Aouf}
\vspace{1ex}
\IEEEauthorblockA{\IEEEauthorrefmark{5}Corresponding author}
\vspace{1ex}
\IEEEauthorblockA{\textit{Department of Engineering, School of Science and Technology}\\
City, University of London, ECV1 0HB London, United Kingdom\\
Email:\{mahmoud.abdulsalam, nabil.aouf,\}@city.ac.uk}
}
\maketitle
\thispagestyle{plain}
\pagestyle{plain}

\begin{abstract}
As demand for robotics manipulation application increases, accurate vision based 6D pose estimation becomes essential for autonomous operations. Convolutional Neural Networks (CNNs) based approaches for pose estimation have been previously introduced. However, the quest for better performance still persists especially for accurate robotics manipulation. This quest extends to the Agri-robotics domain.  In this paper, we propose TransPose, an improved Transformer-based 6D pose estimation with a depth refinement module. The architecture takes in only an RGB image as input with no additional supplementing modalities such as depth or thermal images. The architecture encompasses an innovative lighter depth estimation network that estimates depth from an RGB image using feature pyramid with an up-sampling method. A transformer-based detection network with additional prediction heads is proposed to directly regress the object's centre and predict the 6D pose of the target. A novel depth refinement module is then used alongside the predicted centers, 6D poses and depth patches to refine the accuracy of the estimated 6D pose. We extensively compared our results with other state-of-the-art methods and analysed our results for fruit picking applications. The results we achieved show that our proposed technique outperforms the other methods available in the literature. 
\end{abstract}


\section{Introduction}
6D object pose estimation is a crucial topic to address in the robotics domain. The ability to perceive the position of an object from a single RGB-image can find application in areas such as: robotics for grasping tasks \cite{zhu2014single}, autonomous driving \cite{menze2015object}, space applications \cite{rondao2018multi} and  robotics for virtual and augmented reality applications \cite{marchand2015pose}. This problem, however, comes with several challenges such as: object appearance and texture, lighting conditions and object occlusion \cite{xiang2017posecnn}.
Conventionally, 6D object pose estimation problem is formulated as a feature mapping problem where feature points of a 3D objects are matched on 2D images \cite{lowe1999object,collet2011moped,rothganger20033d}. However, these methods are unable to detect features on smooth objects with minimum or no texture. Introduction of additional modality such as depth data have been used to solve the problem of features on texture-less objects \cite{bo2014learning,hinterstoisser2012model,brachmann2014learning}. However, this requires more inputs in the form of RGB-D images. With the emergence of Convolutional Neural Networks (CNNs), some research leveraged on this powerful tool as part of their pipeline to estimate 6D poses \cite{xiang2017posecnn, wang2019densefusion}. Transformer based models are emerging and proving to be more efficient than CNNs \cite{carion2020end,dosovitskiy2020image, khan2022transformers, touvron2021training}. Thus, few pipelines adopting transformer based models for 6D pose estimation in quest for better accuracy \cite{amini2021t6d,jantospoet,beedu2022video} exist.
In this work, we propose a new 6D object pose estimation architecture where we aim at improving the accuracy in comparison of the existing methods. We introduce TransPose: an improved transformer-based 6D pose estimation network with a novel depth refinement module. The objective is to get better 3D translations and rotations estimates from a single RGB image input. For our initial estimations, we adapted the Detection Transformer (DETR) framework, \cite{carion2020end}, to directly regress the center of the target object. Furthermore, we obtain an image patch of the target object. The translation and rotation can directly be regressed by formulating additional prediction heads on DETR \cite{amini2021t6d}. Indeed, feed-forward heads are added to regress the two components of the 6D pose (3D translation and 3D rotation). A novel depth refinement module is also introduced in our estimation pipeline to increase the accuracy of the pose estimation.

\begin{figure*}[h]
\centering
  \includegraphics[width=0.9\textwidth,height=5cm]{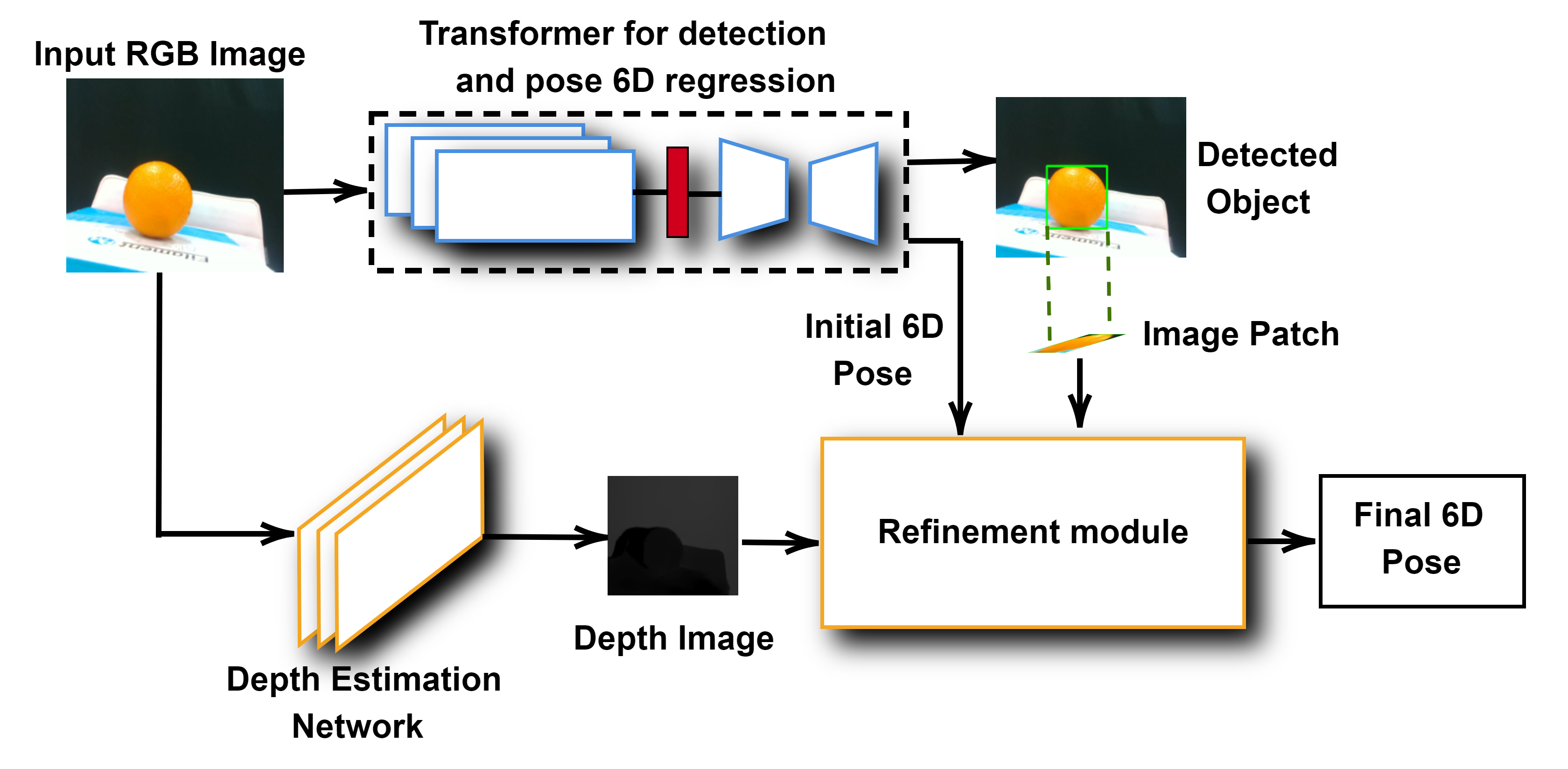}
  \caption{Overall solution architecture performing object detection, depth prediction and 6D pose prediction.}\label{overview}
\end{figure*}

TransPose architecture performs two interdependent tasks to obtain the final 6D pose of the target object. As seen in Fig. \ref{overview}, an RGB image is used as the input to the pipeline. The image is passed to the transformer network which has a ResNet-101 \cite{he2016deep} backbone for features extraction. These features are then passed to the transformer model consisting of a standard encoder and decoder setup \cite{vaswani2017attention}. The model is used to obtain an image patch by detecting the object and assigning a Region Of Interest (ROI) to the detected object. The second segment of the architecture is the depth estimation and refinement module. The depth estimation network encompasses a feature pyramid network (FPN) \cite{lin2017feature} that takes in an RGB image as input and outputs an estimated depth image. The image patch obtained from the transformer model is used to isolate the target on the depth image and hence obtain the depth of the target from the camera. The depth is then used to compute other components of the translation and subsequently used to refine the estimated 6D pose of the target. We evaluated our approach on YCB-Video dataset \cite{xiang2017posecnn} as a benchmark and compared it with other state-of-the-art approaches. The following are our contributions in the TransPose model:
\begin{itemize}
    \item  We propose a novel pipeline for 6D object pose prediction that favourably compares with other state-of-the-art methods
    \item As part of the pipeline, we propose a lighter depth estimation network that utilizes a better up-sampling method for depth prediction
    \item Additional analyses are conducted with our own generated fruit dataset to facilitate and evaluate 6D pose estimation performance for fruit-picking applications. 
    \end{itemize}

The paper continues with a literature review in section II. After introducing a TransPose solution for 6D pose estimation in section III, we provide our results in the experiments section IV and finally the conclusion. 

\section{Related Work}
Many methods have been proposed to tackle the problem of 6D object pose estimation. Approaches that are non-learning-based rely heavily on object textures for pose estimation. Scale-Invariant Feature Transform (SIFT) features \cite{lowe2004distinctive} and Speeded Up Robust Features (SURF) \cite{bay2008speeded} are common examples of the classical features used. The SIFT algorithm as used in \cite{zhang2014sift} for pose estimation requires rich texture information. This can be an issue if the objects are textureless objects.  Miyake et al. \cite{miyake20203d} compensated the textureless nature of objects with the colour information to improve the accuracy of the 6D pose estimation. The geometric information has also been used to increase the accuracy of estimation \cite{zhang2018vision}.

Pose estimation methods that utilise local descriptors define and compute the global descriptors offline. The local descriptor is then computed and matched online with the global descriptor. Pose estimation using Iterative Closest Point (ICP), Oriented Fast and Rotated Brief (ORB) \cite{rublee2011orb}, Binary Robust Independent Elementary Features (BRIEF) \cite{calonder2010brief} have been implemented in the past \cite{guo2019fast,akizuki2018pose,yu2018robust}. However, these methods are computationally expensive and do not perform well on reflective objects.

We can further group pose estimation methods into template-based and featured-based methods \cite{xiang2017posecnn}. The advantage of the template-based methods is that they can detect objects without enough textures. Each input image location is scanned and matched with a constructed template of the object. The best match is selected based on a similarity score that compares the matched locations \cite{hinterstoisser2011gradient,cao2016real,hinterstoisser2012model}. This type of method cannot properly estimate occluded objects since the similarity score will be low.

The feature-based methods utilize 2D-3D correspondences. Features are mapped from the 2D images to the 3D models thereby estimating the 6D poses \cite{tulsiani2015viewpoints,rothganger20033d,lowe1999object}. This approach handles occluded objects better. However, this is at the expense of rich features in the form of enough texture. Some works have proposed learning feature descriptors to solve the problem of objects with no texture \cite{doumanoglou2016siamese, wohlhart2015learning}, while others regress directly from the 2D image to obtain the 3D correspondence \cite{amini2021t6d, krull2015learning, brachmann2014learning, brachmann2016uncertainty}. Without sufficient refinement, these models can obtain relatively low accuracy when dealing with symmetrical objects.

Convolutional Neural Network (CNN) architecture for pose estimation was introduced by \cite{kendall2015posenet} to regress 6D pose using RGB image. Limited by depth modality, the task becomes difficult. In an attempt to address this problem, another method proposed the prediction of depth from the 2D image and thus acquire the 3D position of the object \cite{xiang2017posecnn}. Estimating the rotation component can also be a problem using this method due to non-linearity. \cite{su2015render,sundermeyer2018implicit,liu2016ssd} separated the rotation component and treated it as a classification problem. This often requires a post-refinement to obtain an accurate estimation. Methods that detect keypoints to estimated 6D pose have been proposed to robustly and efficiently estimate 6D pose. \cite{rad2017bb8} utilized a segmentation technique to isolate the Region of Interest (ROI) and further regressed the keypoints from the ROI. Similarly, \cite{tekin2018real} utilised the YOLO \cite{redmon2017yolo9000} framework for such. However, these methods in the face of occlusion perform poorly. To address this problem, some methods obtain keypoints through pixel-wise heatmaps \cite{oberweger2018making,pavlakos20176}. Considering that heatmaps are fixed-size, these methods suffer when the objects are truncated. 

Some other methods have considered using models encompassing classical algorithm such as PnP algorithm to increase the accuracy of estimation \cite{rad2017bb8,peng2019pvnet,hu2019segmentation}. Such models are weighty and hence not always suitable for real-time platform deployment. Models such as the PoseCNN \cite{xiang2017posecnn} and T6D-direct \cite{amini2021t6d} although are able to regress the 6D poses, however a very large dataset is required to train those models since they have no refinement module to count on.  
\begin{figure*}[h]
  \includegraphics[width=1\textwidth,height=6cm]{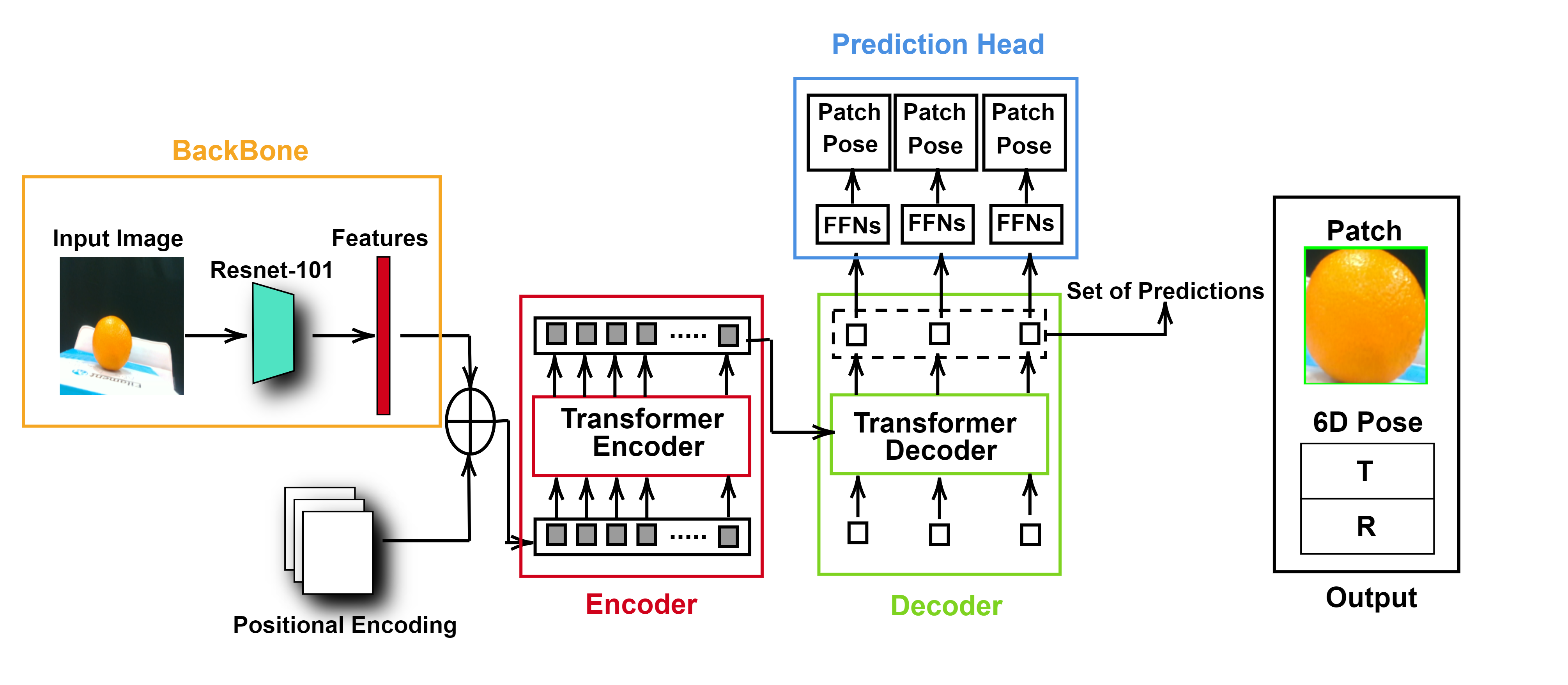}
  \caption{Transformer for detection, image patch and initial 6D pose regression.}\label{xfmer}
\end{figure*}
Pose estimation using depth modality often involve the conversion of depth image to point cloud and proceeds with the segmentation of object masks  \cite{gao2021cloudaae, gao20206d,liu2022catre} adopted semantic segmentation from depth images and point clouds to regress 6D poses. This is accompanied by computational burden due to the conversion to point cloud and often requires a large dataset. In contrast, we utilised the raw depth modality for the regressed pose refinement without converting to point cloud as presented further in this paper. 
\section{TransPose}
The pipeline for TransPose 6D object pose estimation solution we propose in this work can be divided into three main parts:
\begin{itemize}
    \item  Detection and Regression Transformer 
    \item  Depth Estimation Network (DEN)
    \item  Refinement Module for Final 6D Pose Estimation.
    \end{itemize}
\subsection{Detection and Regression Transformer}
This transformer network is mainly adopted for object detection, image patch designation and initial 6D pose regression. The transformer architecture is inspired from Detection Transformer DETR \cite{carion2020end} and T6D-Direct \cite{amini2021t6d}. Our model is presented in Fig. \ref{xfmer}. An RGB image is used as the input of the model. A ResNet-101 is adopted as the CNN backbone to extract and create a feature vector which is used as an input to the transformer encoder-decoder. Set of predictions of size $N_c$ are produced by the transformer encoder-decoder. Prediction heads are added in form of Feed Foward Networks (FFNs) to regress the object pose and patch.  The losses adopted to train this transformer network are categorized as follows:
\subsubsection{Set Prediction Loss}
The patch prediction in form of ROI is obtained by assigning a bounding box around the object of interest. From the input image through the decoder, the model produces a set of size $N_{c}$ of tuples with fixed cardinality, where $N_{c}$ also corresponds to the maximum number of the expected targets within the image. The content of each tuple is the image patch (left bottom pixel coordinates, height and width), class label probabilities and 6D pose (translation and rotation) of the predicted object.  A bipartite matching is adopted to match the ground truth and the predicted sets to obtain matching pairs. The model is then trained to minimise a loss between the pairs. 

Consider ground truth objects $x_{1}$, $x_{2}$, $x_{3}$, ... $x_{n}$, let's assume $N_{c}$ is more than the number of objects in the image, bipartite matching is performed to match the ground truth $x$ which is a set of size $N_{c}$ padded with no-object ($\emptyset$) with the predicted set $\hat{x}$ of the same size. Essentially, performing a permutation between the sets while minimizing the loss below.
\begin{equation}
\hat{\rho} = \underset{\rho \in \Theta_{N_{c}}}{\arg\min}\sum_{i} ^{N_{c}} \mathcal{L}_{match}(x_{i}, \hat{x}_{\rho (i)})\label{1}
\end{equation}

$\mathcal{L}_{match}(x_{i}, \hat{x}_{\rho (i)})$ is the pair-wise match cost between the prediction at index $\rho (i)$ and the ground truth tuple $x_{i}$.

\subsubsection{Hungarian loss}
After matching, the model is trained to minimise the Hungarian loss. We denote the predicted patch as $\hat{\gamma}_{\rho (i)}$. Thus, the hungarian loss is defined as below:

\begin{equation}
\begin{array}{ll}
 \mathcal{L}_{hung}(x_{i}, \hat{x})=& \sum\limits_{i} ^{N_{c}}[\lambda _{pose}\mathbbm{1}_{c_{i}\neq\emptyset}\mathcal{L}_{pose}(R_{i},t_{i},\hat{R}_{\hat{\rho} (i)}, \hat{t}_{\hat{\rho} (i)}) ~~\\ & -log\hat{P}_{\rho (i)}(c_{i})  +
 \mathbbm{1}_{c_{i}\neq\emptyset}\mathcal{L}_{patch}(\gamma_{i}, \hat{\gamma}_{\hat{\rho} (i)})]
 
 \end{array}
 \label{hungarian}
\end{equation}

$\hat{\rho}$ is the lowest cost from equation.\ref{1}, $c_i$ is the class probability and $\gamma _i$ is a vector that defines the ground truth image patch coordinates, height and width. 
\subsubsection{Patch loss}
The patch loss $\mathcal{L}_{patch}(\gamma_{i}, \hat{\gamma}_{\rho (i)})$ is a component of equation \ref{hungarian} and combines an $l_1$ norm loss and a generalized loss
$\mathcal{L}_{iou}(\gamma_{i}, \hat{\gamma}_{\rho (i)})$,
\cite{rezatofighi2019generalized}, as follows:
\begin{equation}
\mathcal{L}_{patch}(\gamma_{i}, \hat{\gamma}_{\rho (i)}) = \sigma _{1}\mathcal{L}_{iou}(\gamma_{i}, \hat{\gamma}_{{\rho} (i)}) +  \sigma _{2} \lvert \lvert \gamma_{i} - \hat{\gamma}_{\rho (i)} \rvert \rvert \label{patch}
\end{equation}

and,
\begin{equation}
\mathcal{L}_{iou}(\gamma_{i}, \hat{\gamma}_{\rho (i)}) = 1 - \left(\frac{\lvert (\gamma_{i} \cap \hat{\gamma}_{\rho (i)} \rvert} {\lvert (\gamma_{i} \cup \hat{\gamma}_{\rho (i)} \rvert} - \frac{\lvert L(\gamma_{i}, \hat{\gamma}_{\rho (i)}) \backslash \gamma_{i} \cup \hat{\gamma}_{\rho (i)} \rvert }{\lvert L(\gamma_{i}, \hat{\gamma}_{\rho (i)})  \rvert}\right)
\end{equation}

$\sigma _{1}$,$\sigma _{2}$ $\in$ $\mathbb{R}$ are hyperprameters.  $L(\gamma_{i}, \hat{\gamma}_{\rho (i)})$ is the largest patch having the ground truth $\gamma_{i}$ and the predicted $\hat{\gamma}_{\rho (i)}$.

\subsubsection{Pose loss}
$\mathcal{L}_{pose}(R_{i},t_{i},\hat{R}_{\hat{\rho} (i)}, \hat{t}_{\hat{\rho} (i)})$ is the pose loss. The pose loss is divided into two components, translation $t$  and the Rotation $R$. Conventional $l_2$ norm loss is used to supervise the translation while a ShapeMatch loss ${L}_{R}$, \cite{xiang2017posecnn}, is used for the rotation to deal with symmetrical objects.
\begin{equation}
\mathcal{L}_{pose}(R_{i},t_{i},\hat{R}_{\rho (i)}, \hat{t}_{\rho (i)}) = {L}_{R}(R_{i}, \hat{R}_{\rho (i)}) + \lvert \lvert t_{i} - \hat{t}_{\rho (i)} \rvert \rvert
\end{equation}

\begin{equation}
{L}_{R} = 
\left\{ \begin{array}{ll}
\frac{1}{\lvert K \rvert} \underset{j_1 \in K}{\sum} \  \underset{j_2 \in K}{\min} \lvert \lvert (R_{i}j_{1} - \hat{R}_{\rho (i)}j_{2}) \rvert \rvert \ & \text{if symmetric,} \\\\
\frac{1}{\lvert K \rvert} \underset{j \in K} {\sum}\lvert \lvert (R_{i}j - \hat{R}_{\rho (i)}j) \rvert \rvert \ & ~ \text{otherwise.}
\end{array} \right.
\end{equation}

$K$ represents the 3D points set. $R_{i}$and $t_{i}$ are the ground truth rotation and translation, respectively. $\hat{R}_{\rho (i)}$ and $\hat{t}_{\rho (i)}$ are the respective predicted object rotation and translation. 

\subsection{Depth Estimation Network (DEN)}
Depth estimation can be used for many applications \cite{araar2015power}. In our case, the DEN is responsible for estimating depth images from monocular images inspired by the Feature Pyramid Network (FPN) \cite{lin2017feature}. The motivation is that FPN is capable of extracting features at different scales. We adopt ResNet-101 network as a backbone for feature extraction, where two 3x3 convolutional layers are utilised to process features and ReLU as an activation function for the layers as seen in Fig. \ref{depthy}. A better lightweight upsampling technique \cite{wang2019carafe} that covers a larger field of view and enables the generation of adaptive kernels for better prediction is utilised. The depth images are one-fourth of the original image's size. The gradient of the depth map is obtained using a Sobel filter.  The depth loss adopted in the training of our network is an $l_1$ norm loss defined as follows:

\begin{equation}
\mathcal{L}_{depth} = \frac{1}{n}\sum _{i=1} ^ n \lvert \lvert d_i - \hat{d} _{(i)} \rvert \rvert
\end{equation}

where, $d_i$ and $\hat{d} _{(i)}$ are the ground truth depth and the predicted depth of every pixel $i$ respectively.


\subsection{Refinement Module for Final 6D Pose Estimation.}\label{SCM}
\begin{figure*}[h!]
\centering
  \includegraphics[width=0.9\textwidth,height=5cm]{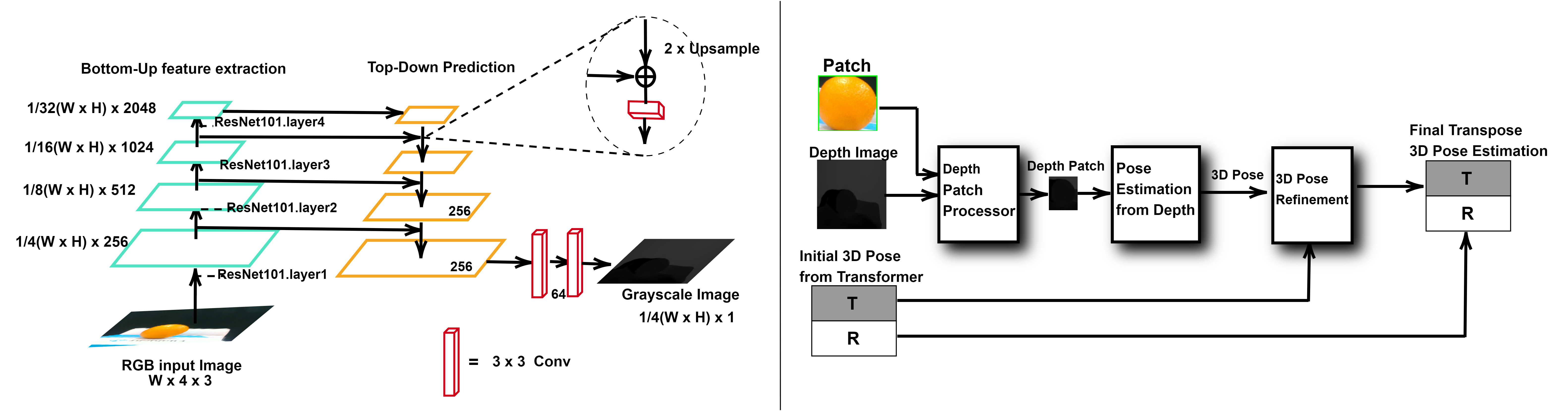}
  \caption{(a) Depth estimation network using feature pyramid ~~(b) Refinement module for final 3D pose estimation}\label{depthy}
\end{figure*}

The refinement module consists of the depth patch generation and final pose estimation processes. The patch and the regressed 6D pose from the transformer alongside the depth image are used as inputs for the refinement module as shown in Fig. \ref{depthy}.

The patch defined as the ROI obtained  by the Detection and Regression Transformer is formulated as:

\begin{equation}
\psi _{i} = [B_{opx}, B_{opy}, H_{op}, W_{op}] \label{psi}
\end{equation}
where $B_{opx}, B_{opy}$ represent the bottom left corner pixel coordinates of the patch respectively and $H_{op}, W_{op}$ are the height and width of the patch respectively, all with respect to the original RGB image size (height and width) $S_o$ = $(W_o \times H_o)$. Similarly, let us represent the size of the depth image as $S_d$ = $(W_d \times H_d)$. where $S_o \neq S_d$. We can obtain our depth patch $\psi _j$ with respect to $S_d$ from equ. \ref{psi} as:

\begin{equation}
\begin{aligned}
\psi _{j} = [B_{dpx}, B_{dpy}, H_{dp}, W_{dp}] \\
\\
 = \psi _{i} \times 
\begin{bmatrix}
\frac{W_d}{W_o} & 0 & 0 & 0\\
0 & \frac{H_d}{H_o} & c & 0\\
0 & 0 & \frac{H_d}{H_o}  & 0\\
0 & 0 & 0 & \frac{W_d}{W_o}\\
\end{bmatrix}
\end{aligned}
\end{equation}
where $B_{dpx}, B_{dpy}$ now represents the bottom left pixel coordinates of the depth patch respectively and $H_{dp}, W_{dp}$ are the height and width of the depth patch respectively, all with respect to the depth image size $S_d$. The depth patch represents now our object ROI in the depth image frame and thus we can obtain the depth $t_{z1}$ from the camera to the target to be the depth information at the center pixel of the depth patch. The center pixel coordinates $C_d = (C_{dx}, C_{dy})^T$ can be obtained as follows:

\begin{equation}
\begin{aligned}
C_{dx} = B_{dpx} + \frac{W_{dp}}{2}\\
C_{dy} = B_{dpy} + \frac{H_{dp}}{2}
\end{aligned}
\label{cen}
\end{equation}

The translation from the depth network model $t_1$ utilises $t_{z1}$ (which in this case is the depth) to compute $t_{x1}$ and $t_{y1}$ which are the translations in $x$ and $y$ axis to complete the translation $t_1 = (t_{x1},t_{y1},t_{z1})^T$. Assuming the camera matrix is known, $t_{x1}$ and $t_{y1}$ can be obtained following the projection equation of a pinhole camera model as follows:
\begin{equation}
\begin{bmatrix}
C_{ox}\\
C_{oy}\\
\end{bmatrix}
=
\begin{bmatrix}
f_x\frac{t_{x1}}{t_{z1}} + PP_x\\
f_y\frac{t_{y1}}{t_{z1}} + PP_y\\
\end{bmatrix}
\end{equation}

where $f_x$ and $f_y$ represent the focal length of the camera, $( PP_x, PP_y)^T$ is the principal point. $C_o = (C_{ox}, C_{oy})^T $ is the object centroid, which can be obtained from the image patch similarly to equ. \ref{cen} to be $(B_{opx} + \frac{W_{op}}{2}, B_{opy} + \frac{H_{op}}{2})^T$ assuming the centroid coincides with the center of the patch. Thus, $t_{x1}$ and $t_{y1}$ can be calculated as:
\begin{equation}
\begin{bmatrix}
t_{x1}\\
t_{y1}\\
\end{bmatrix}
=
\begin{bmatrix}
\frac{(C_{ox}- PP_x)t_{z1}}{f_x}\\
\frac{(C_{oy}- PP_y)t_{z1}}{f_y}\\ 
\end{bmatrix} \label{12}
\end{equation}
Thus a complete translation from the depth image $t_1$ is obtained as: 
\begin{equation}
   t_1 = (t_{x1}, t_{y1}, t_{z1})^T \label{13}
\end{equation}

Finally, we can obtain the final fusion-based object translation $t$ as:
\begin{equation}
t = (w_1 \times t_1) + (w_2 \times t_2)
\end{equation}

where the weights $w_1,w_2 \geq 0$ and $w_1+w_2 = 1$. $t_1$ is the computed translation from the depth in equ. \ref{13} and $t_2$ is the regressed translation from the transformer model. Note that $w_1$ and $w_2$ are selected depending on the performance of both the transformer and depth model. Such that, the model with a lower loss will have a higher $w$ and vice-versa. 

\section{Experiments}
In the following, we present all the experiments conducted to test the capability of our proposed TransPose solution. From the datasets adopted to the results and comparison made between our solution and existing solutions, all will be detailed in the following subsections. 

\subsection{Dataset}
The popular KITTI dataset is used as a benchmarking dataset for the depth estimation network. Likewise, we use the popular YCB-Video dataset being a benchmark for 6D pose estimation. \cite{xiang2017posecnn} so we can easily compare our results with other methods. The dataset has 133,936 images of 640 $\times$ 480 pixels resolution. Each image is accompanied with bounding box labels, depths, segmentation and 6D object pose annotations. Similar to \cite{xiang2017posecnn}, a test was carried out on 2,949 keyframes from 12 scenes. Additionally, we sampled from the Fruity dataset \cite{} to validate this approach in the context of fruit picking application which is an important application for our research.
\subsection{Evaluation Metrics}
The metrics adopted to evaluate the depth estimation network are the $abs$-$rel$, $sq$-$rel$,  $RMSE$ and $RMSE_{log}$, as proposed in \cite{eigen2014depth}, as follows:
\begin{eqnarray}
abs_{-}rel &=& \frac{1}{\lvert T \rvert}\sum _{i = 1}^T  \frac{ \lvert d_i - \hat{d} _{i} \rvert }{\hat{d} _{i}} \label{abs}\\
sq_{-}rel &=& \frac{1}{\lvert T \rvert}\sum _{i = 1}^T  \frac{\lvert \lvert d_i - \hat{d} _{i} \rvert \rvert^2 }{\hat{d} _{i}} \label{sq}\\
RMSE &=& \sqrt{\frac{1}{\lvert T \rvert}\sum _{i = 1}  \lvert \lvert d_i - \hat{d} _{i} \rvert \rvert^2} \label{rmse} \\
RMSE_{log} &=& \sqrt{\frac{1}{\lvert T \rvert}\sum _{i = 1}  \lvert \lvert  \log d_i - \log \hat{d} _{i} \rvert \rvert^2} \label{rmse_log}
\end{eqnarray}
where $T$ is the number of pixels in the test set. 

For the evaluation of the overall pose estimation, the average distance (ADD) metric, as suggested in \cite{pavlakos20176}, is used. This metric calculates the mean pairwise distance as follows:

\begin{equation}
ADD = \frac{1}{\lvert K \rvert}\sum_{j \in K}\lvert \lvert (R_j + t) - (\hat{R}_j + \hat{t}) \rvert \rvert \label{add}
\end{equation}
where $R$ and $t$ are the ground truth object rotation and translation, respectively. $\hat{R}$ and $\hat{t}$ are the predicted rotation and translation respectively. $K$ is the set of 3D points. 

ADD is calculated as the closest point distance for symmetrical objects as follows:
\begin{equation}
ADD-S = \frac{1}{\lvert K \rvert}\sum_{j_1 \in K}\underset{j_2 \in K}{\min} \lvert \lvert (R_{j1} + t) - (\hat{R}_{j2} + \hat{t}) \rvert \rvert \label{add-s}
\end{equation}

\subsection{Training}
The model is initialised as in \cite{carion2020end} with pre-trained weights. The model utilizes an input of image sized 640 $\times$ 480. The initial learning rate is set to $1.0^{-3}$ which is eventually decayed. The batch size is set to 16 samples. AdamW optimizer \cite{loshchilov2017decoupled} is used for the training. The hyperparameters for calculating $\mathcal{L}_{patch}$ in equation. \ref{patch}, $\sigma _1$ and $\sigma _2$ are set to 2 and 5. Also, the parameter $\lambda _{pose}$ for calculating $\mathcal{L}_{hungarian}$ in equation. \ref{hungarian} is set to 0.05. The cardinality or number of prediction queries $N_c$ is set to 21. 

\subsection{Results}
\subsubsection{Depth estimation results}
\begin{figure}[h!]
  \includegraphics[width=0.5\textwidth,height=4cm]{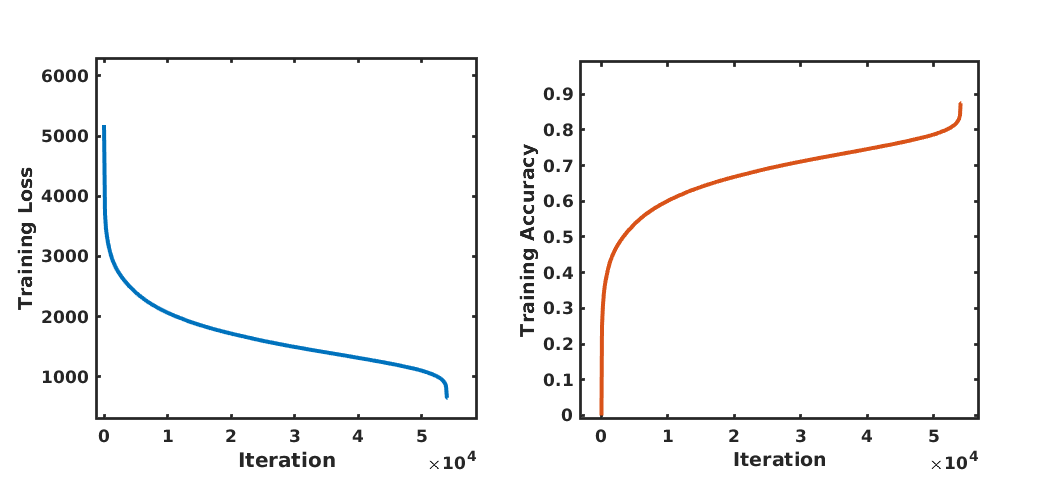}
  \caption{Training loss and accuracy per iteration}\label{training_loss}
\end{figure}

For the depth estimation network, the training loss and accuracy per iteration are shown in Fig. \ref{training_loss}. As the training proceeds, the training loss decreases thereby increasing the training accuracy per iteration.

The results obtained for the depth evaluation using the metrics in equations. \ref{abs}, \ref{sq},\ref{rmse} and \ref{rmse_log} are presented in Table. \ref{tab1}.

\begin{table}[h!]
\caption{Depth estimation Network comparison with other methods}
\begin{center}
\begin{tabular}{|c|c|c|c|c|}
\hline
\textbf{}&\multicolumn{4}{|c|}{\textbf{Evaluation Metric (lower is better)}} \\
\cline{2-5} 
\textbf{Method}&\multicolumn{4}{|c|}{\textbf{KITTI Dataset}} \\
\cline{2-5} 
\textbf{} & \textbf{\textit{$abs$-$rel$}} & \textbf{\textit{$sq$-$rel$}} & \textbf{\textit{$RMSE$}}& \textbf{\textit{$RMSE_{log}$}}  \\
\hline
Make3D\cite{saxena2008make3d} & 0.280 & 3.012 & 8.734 & 0.361\\
\hline
 Eigen et al \cite{eigen2014depth} & 0.190 & 1.515 & 7.156 & 0.270\\
 \hline
 Liu et al \cite{liu2015deep} & 0.217 & 1.841 &6.986 & 0.289\\
  \hline
 Kuznietsov et al \cite{kuznietsov2017semi} & \textbf{0.113} & 0.741 & \textbf{4.621} & 0.189\\
  \hline
Ours &\textbf{0.114} & \textbf{0.724} & \textbf{4.694} & \textbf{0.185}\\
\hline
\textbf{}&\multicolumn{4}{|c|}{\textbf{Custom Fruit Dataset}} \\
\hline
 Eigen et al \cite{eigen2014depth} & 0.0885 & 1.3000 & 4.2440 & 0.2115\\
 \hline
 Liu et al \cite{liu2015deep}&  0.0755 & 1.0917  & 3.9290 & 0.1938\\
  \hline
 Kuznietsov et al \cite{kuznietsov2017semi} & 0.0499 & 0.5350 & 2.6907 & 0.1427 \\
  \hline
Ours & \textbf{0.0434} & \textbf{0.5153} & \textbf{2.5013}  &  \textbf{0.1342}\\
\hline

\end{tabular}
\label{tab1}
\end{center}
\end{table}

We compare the performance of our depth estimation network with other methods on the popular KITTI dataset and our custom fruit dataset. On the KITTI dataset, our method outperformed the others in the $sq$-$rel$ and $RMSE_{log}$ metric and compares very closely with \cite{kuznietsov2017semi} in the $abs$-$rel$ and $RMSE$ metric. On the fruit dataset, our network outperforms the other in $abs$-$rel$, $sq$-$rel$, $RMSE_{log}$ metric and compares closely in the $RMSE$ metric. This comparison shows the accuracy of our network as compared with other literature and the flexibility to adapt for depth estimation as part of the TransPose pipeline. It is worth noting that higher depth accuracy comes at a computational cost and the depth estimation network is just one part as a step of the TransPose pipeline. Thus, a reasonable trade-off between computational cost and accuracy is established to satisfy both decent estimation and future real-time implementation. Hence, the depth results are very satisfactory for our purpose. 

\begin{figure}[h]
  \includegraphics[width=0.5\textwidth,height=10cm]{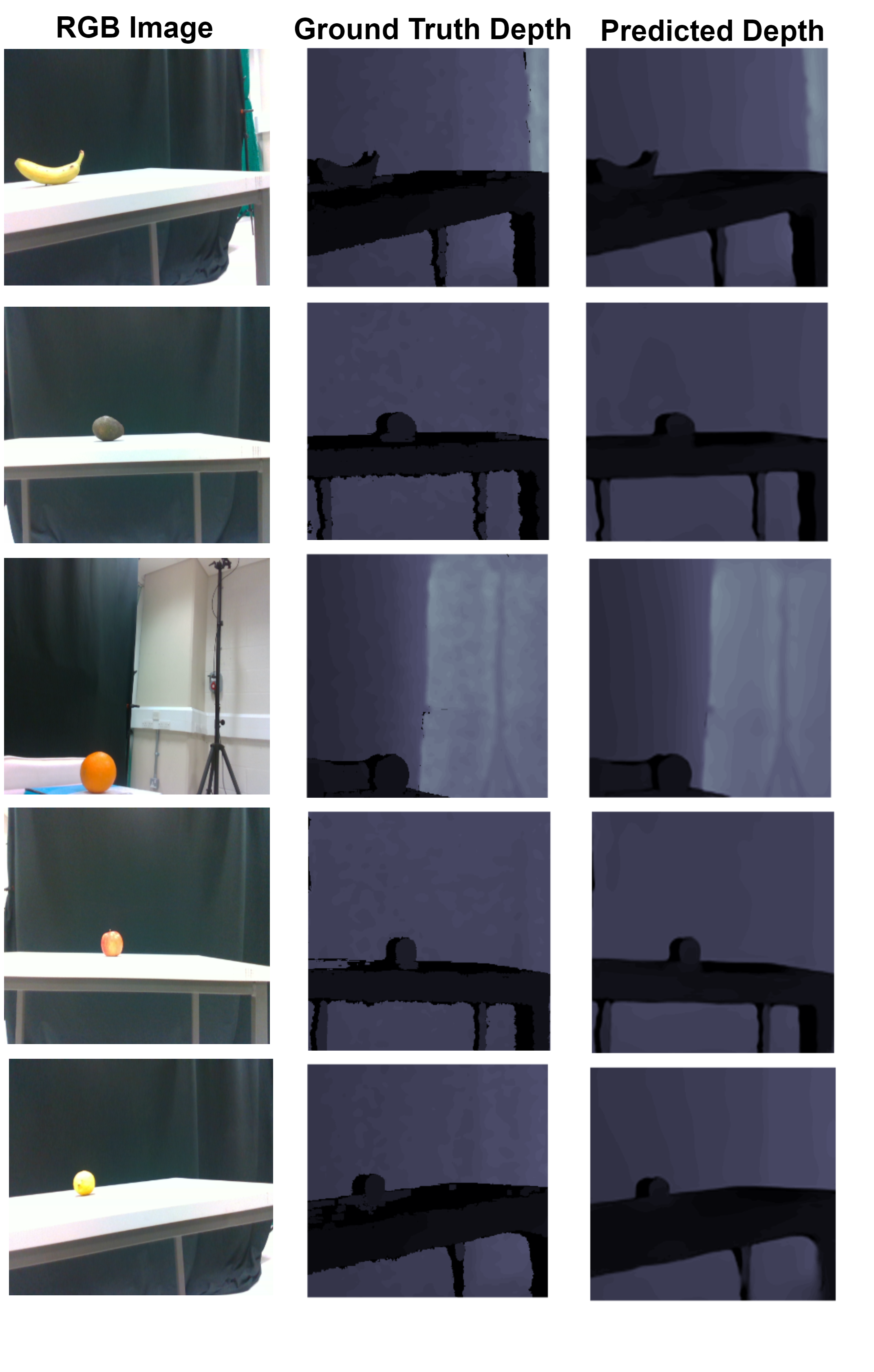}
  \caption{Qualitative results of the depth prediction network. The left column shows the ground truth RGB Images of 5 fruit classes. The middle column shows the ground truth depth images of the corresponding RGB images. The right column shows the predicted depth images from our network. }\label{quali}
\end{figure}

The depth estimation qualitative results are shown in Fig. \ref{quali}. Samples from all the classes of our Fruit dataset including their ground truths and the corresponding predictions are shown. A colour map is added to the depth images for better visualisation and evaluation.  

\begin{figure}[h]
  \includegraphics[width=0.5\textwidth,height=4cm]{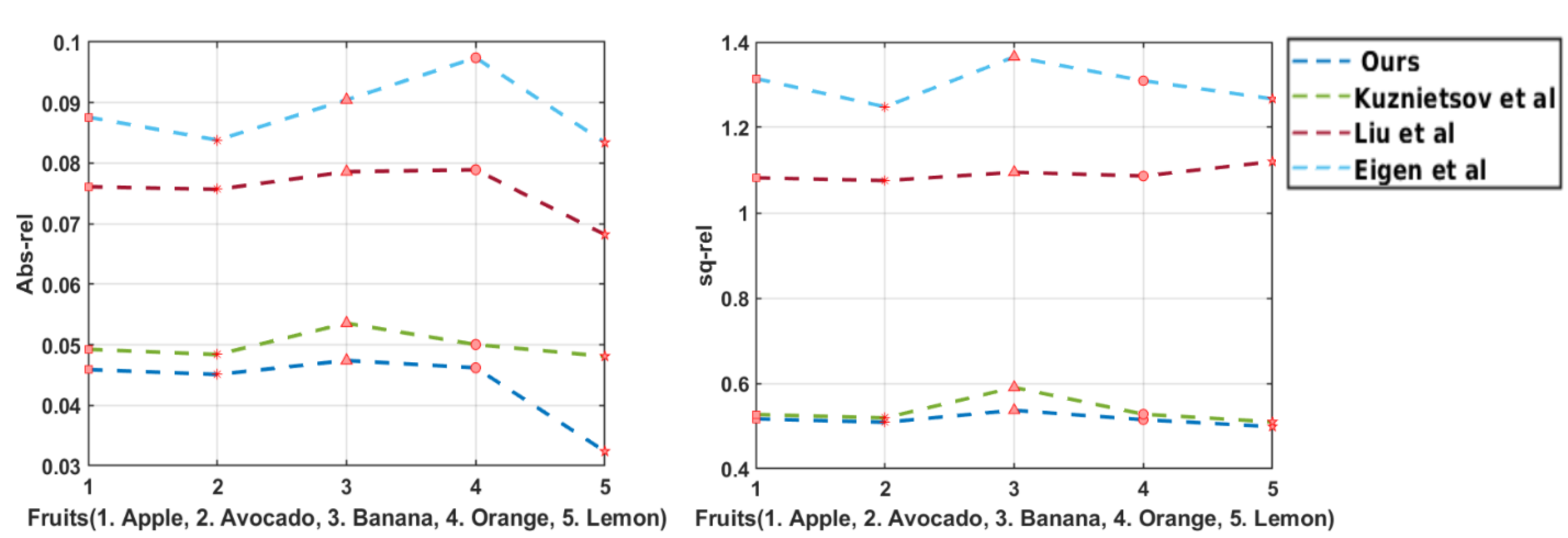}
  \caption{$Abs$-$rel$ and $sq$-$rel$ metric comparison with other methods in the literature. Each fruit class with the corresponding evaluation metric is shown.}\label{comp1}
\end{figure}

Further comparisons with other methods are carried out across each individual class of fruit. Fig \ref{comp1} shows the comparison of each class of the fruit dataset using the $Abs$-$rel$ and $sq$-$rel$ metrics. From the results, our network outperformed all the methods across all the fruit classes. For the $sq$-$rel$, Our depth estimation network performs better in the banana class and slightly performs better in the other fruit classes. 

\begin{figure}[h]
  \includegraphics[width=0.5\textwidth,height=4cm]{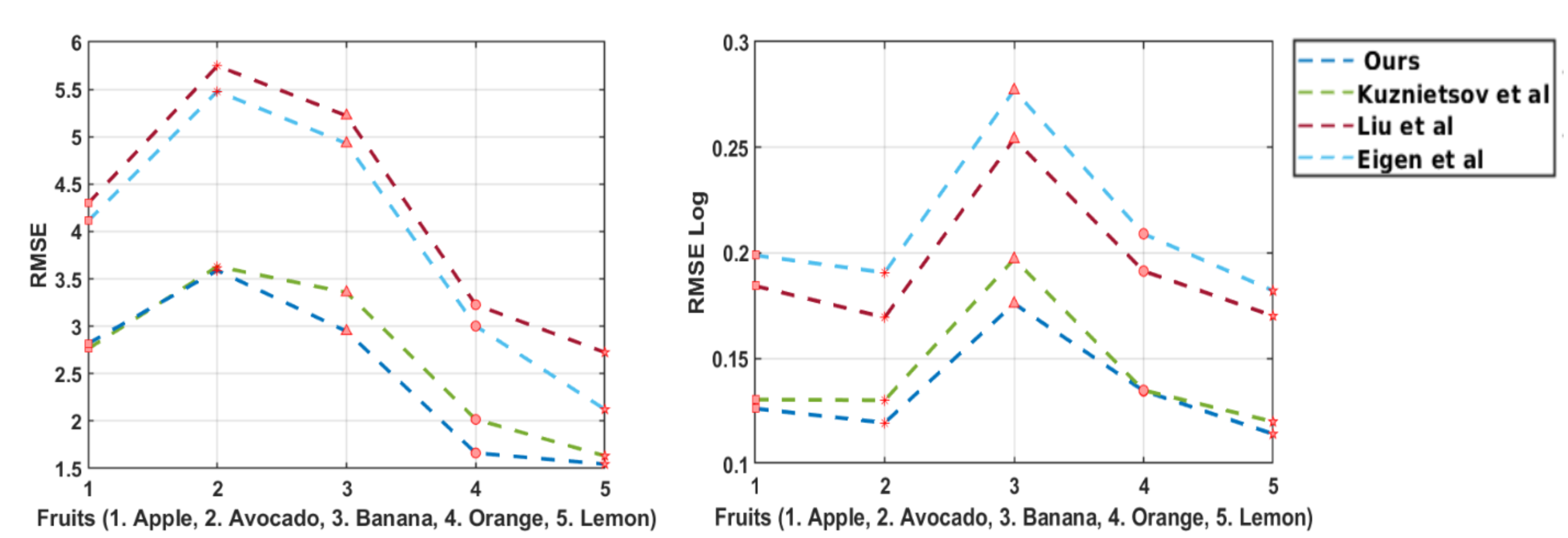}
  \caption{$RMSE$ and $RMSE_{log}$ metric comparison with other methods in literature. Each fruit class with the corresponding evaluation metric is shown.}\label{comp2}
\end{figure}

Fig \ref{comp2} compares the  $RMSE$ and $RMSE_{log}$ of each class of the fruit dataset. Our network performs better on the banana, orange and lemon class using the $RMSE$ metric and compares with \cite{kuznietsov2017semi} on the apple and avocado class. For the $RMSE_{log}$, our network outperforms in the apple, avocado, banana and lemon class. 
\subsubsection{TransPose pose estimation results}
\begin{figure}[h!]
  \includegraphics[width=0.5\textwidth,height=3cm]{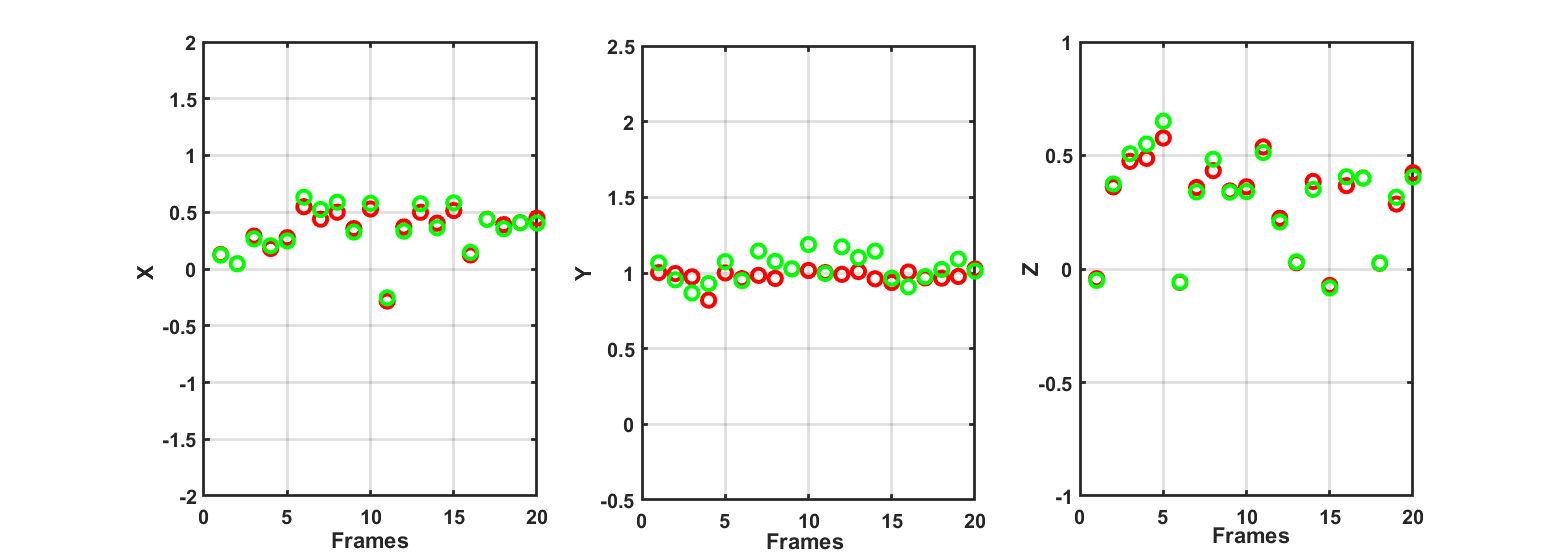}
  \caption{Translation $[t_x, t_y, t_z]^T$ across 20 frames for apple fruit class. Red is the ground truth while green is the prediction.}\label{9}
\end{figure}

\begin{figure}[h!]
\centering
  \includegraphics[width=0.4\textwidth,height=4cm]{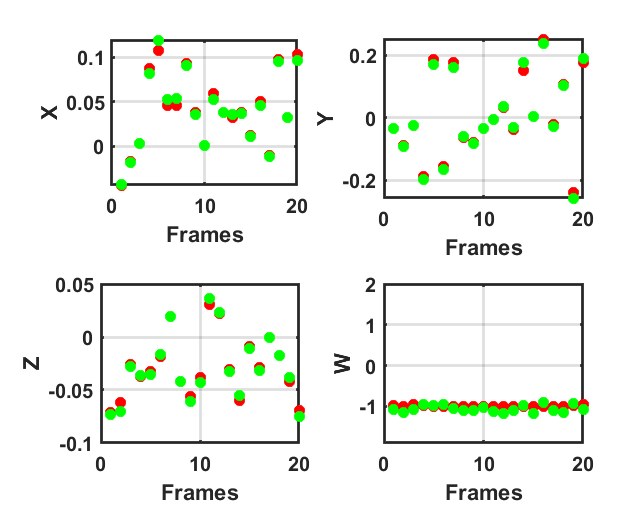}
  \caption{Quaternion $[Q_x, Q_y, Q_z, Q_w]^T$ across 20 frames for orange fruit class. Red is the ground truth while green is the prediction.}\label{18}
\end{figure}

We sample 20 test frames for the 6D pose estimation and compare the ground truth and the predicted poses. The translation $[t_x, t_y, t_z]^T$ and the Quaternion $[Q_x, Q_y, Q_z, Q_w]^T$ which define the orientation are compared for all fruit classes as shown in Fig \ref{9} - \ref{18}.

\begin{figure}[h!]
\centering
  \includegraphics[width=0.4\textwidth,height=4cm]{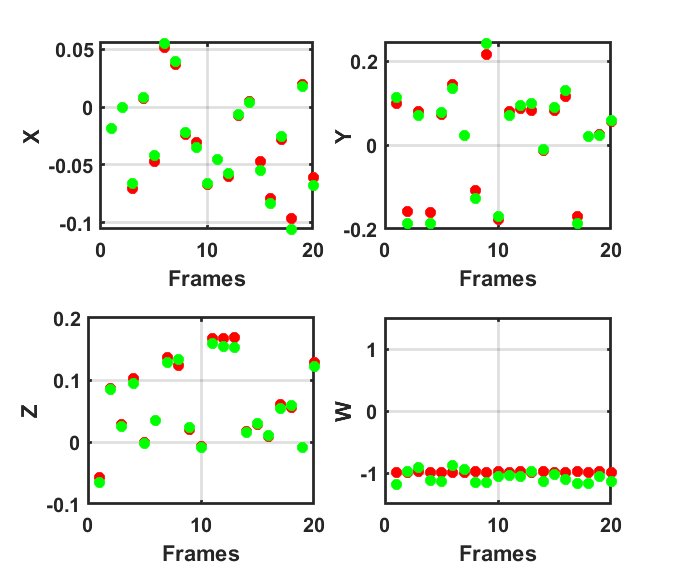}
  \caption{Quaternion $[Q_x, Q_y, Q_z, Q_w]^T$ across 20 frames for apple fruit class. Red is the ground truth while green is the prediction.}\label{refine}
\end{figure}

\begin{figure}[h!]
  \includegraphics[width=0.5\textwidth,height=3cm]{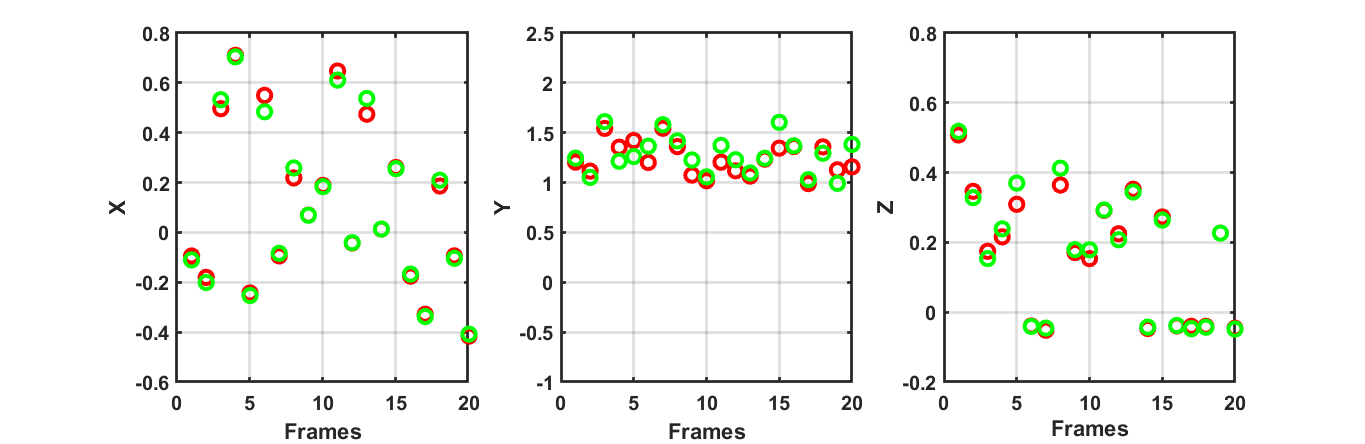}
  \caption{Translation $[t_x, t_y, t_z]^T$ across 20 frames for avocado fruit class. Red is the ground truth while green is the prediction.}\label{refine}
\end{figure}

\begin{figure}[h!]
\centering
  \includegraphics[width=0.4\textwidth,height=4cm]{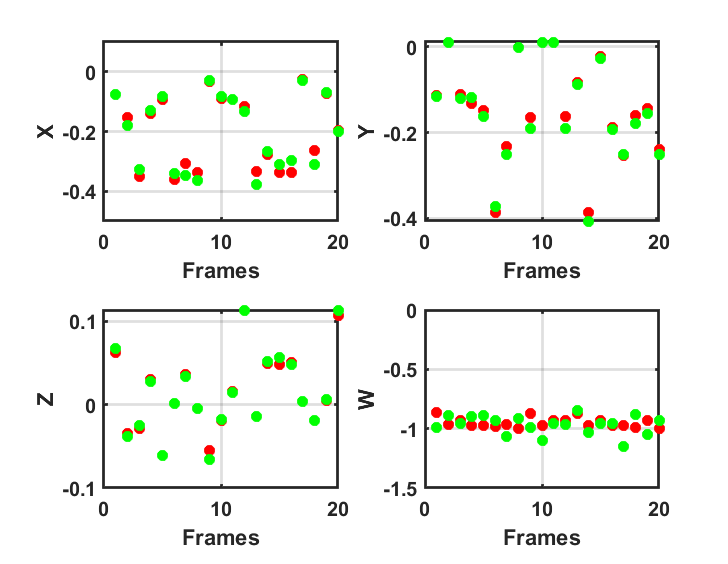}
  \caption{Quaternion $[Q_x, Q_y, Q_z, Q_w]^T$ across 20 frames for avocado fruit class. Red is the ground truth while green is the prediction.}\label{refine}
\end{figure}

\begin{figure}[h!]
  \includegraphics[width=0.5\textwidth,height=3cm]{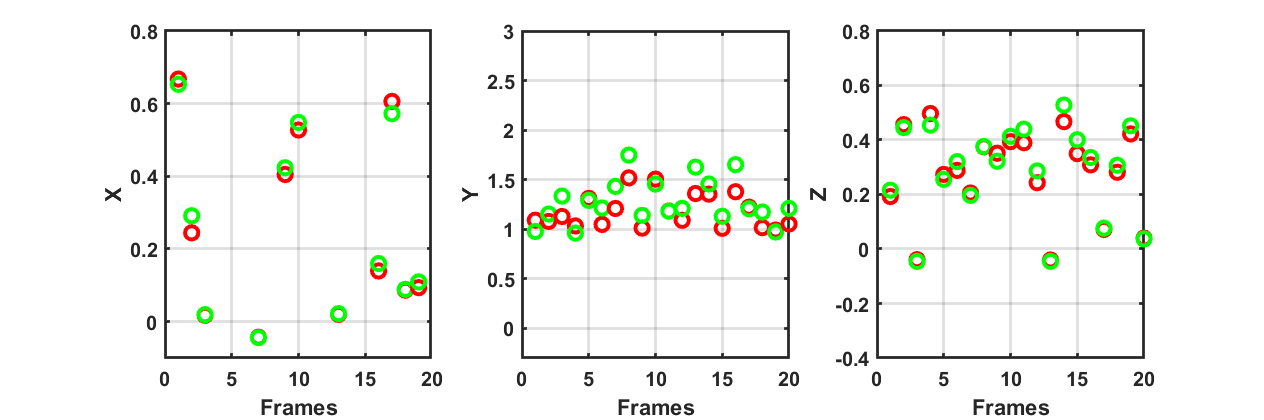}
  \caption{Translation $[t_x, t_y, t_z]^T$ across 20 frames for banana fruit class. Red is the ground truth while green is the prediction.}\label{refine}
\end{figure}

\begin{figure}[h!]
\centering
  \includegraphics[width=0.4\textwidth,height=4cm]{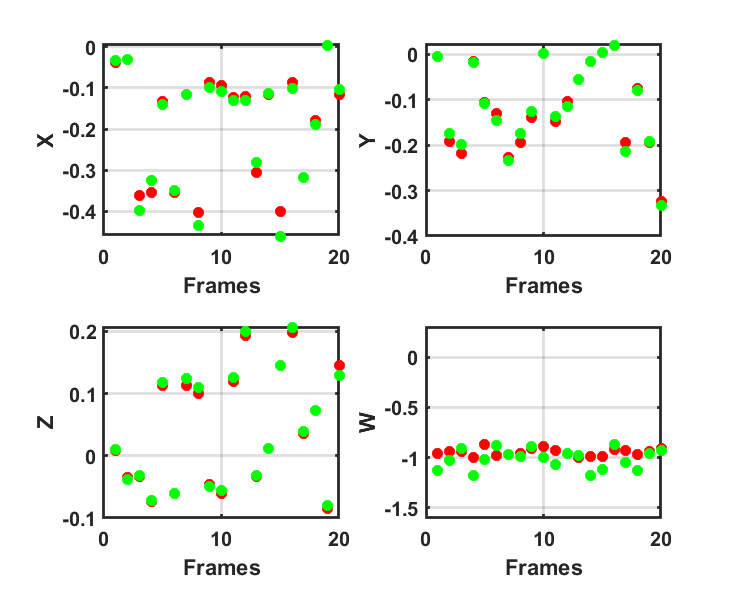}
  \caption{Quaternion $[Q_x, Q_y, Q_z, Q_w]^T$ across 20 frames for banana fruit class. Red is the ground truth while green is the prediction.}\label{refine}
\end{figure}

\begin{figure}[h!]
  \includegraphics[width=0.5\textwidth,height=3cm]{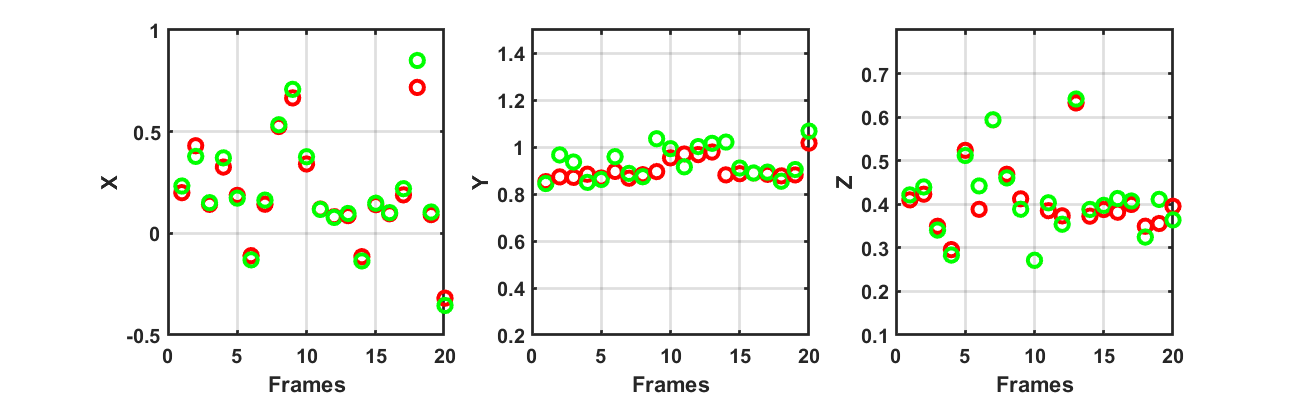}
  \caption{Translation $[t_x, t_y, t_z]^T$ across 20 frames for lemon fruit class. Red is the ground truth while green is the prediction.}\label{refine}
\end{figure}

\begin{figure}[h!]
\centering
  \includegraphics[width=0.4\textwidth,height=4cm]{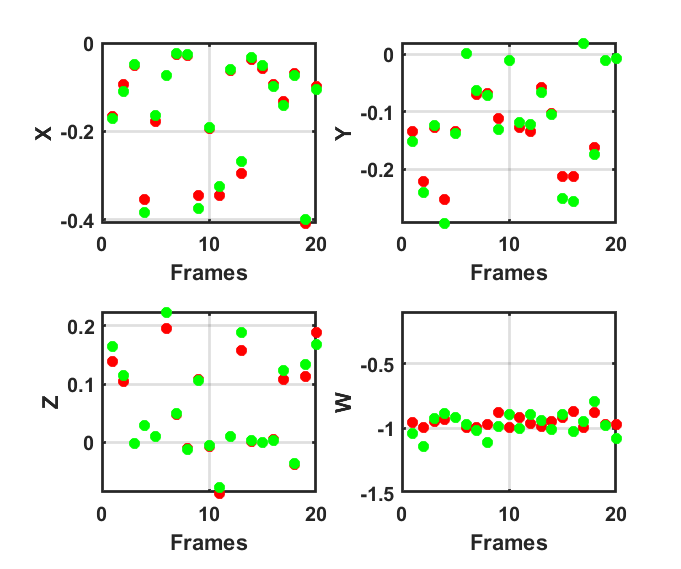}
  \caption{Quaternion $[Q_x, Q_y, Q_z, Q_w]^T$ across 20 frames for lemon fruit class. Red is the ground truth while green is the prediction.}\label{refine}
\end{figure}

\begin{figure}[h!]
  \includegraphics[width=0.5\textwidth,height=3cm]{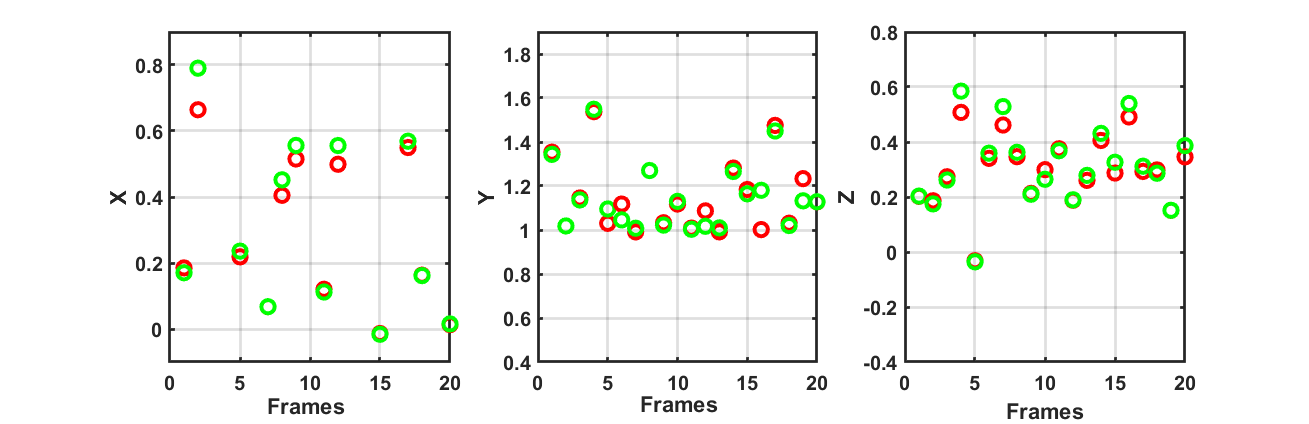}
  \caption{Translation $[t_x, t_y, t_z]^T$ across 20 frames for orange fruit class. Red is the ground truth while green is the prediction.}\label{refine}
\end{figure}

\begin{figure*}[h!]
  \includegraphics[width=\textwidth,height=5cm]{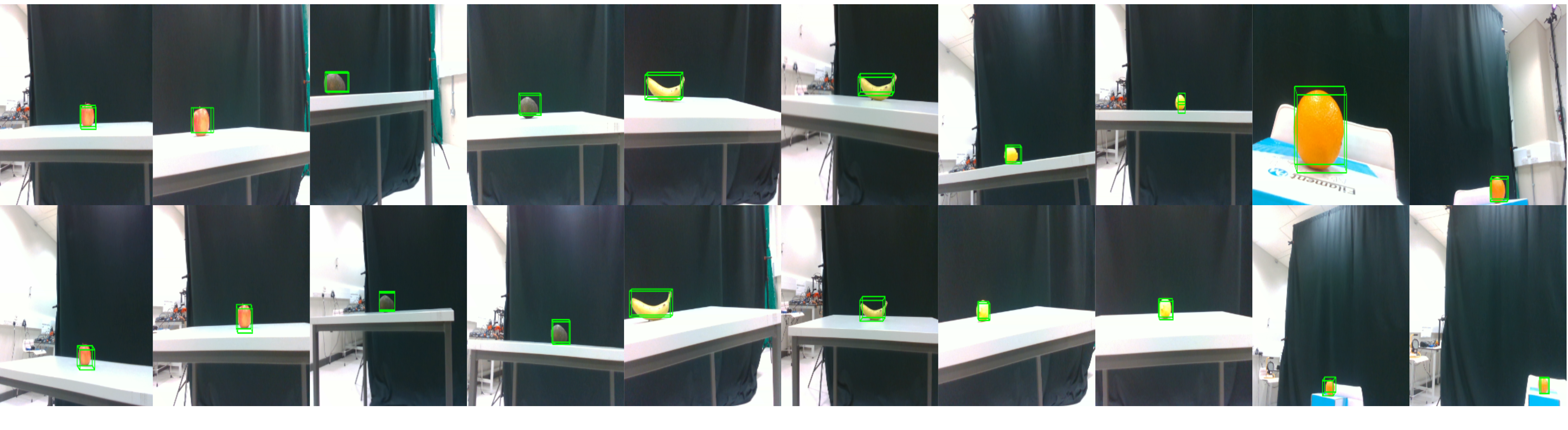}
  \caption{Qualitative samples from the fruit dataset across all the classes}\label{6dquali}
\end{figure*}

The samples are randomly selected from the test data to visualise the difference between the ground truth and the prediction. We can see that our TransPose prediction solution matches well with the ground truth poses across all the fruit classes.
The qualitative results we obtain for some sample frames from the fruit dataset is shown in Fig \ref{6dquali}.

Table \ref{tab2} shows a detailed evaluation of some objects from the YCB dataset  using the metric in equ. \ref{add} and equ. \ref{add-s}.

\begin{table}[h!]
 \sisetup{
 table-number-alignment = center,
 table-figures-integer = 1,
 table-figures-decimal = 4
 }
 \caption{Pose estimation Comparison using various approaches on some objects from YCB-V Dataset. Symmetrical objects are highlighted in green}
 \label{tab2}
 \centering
 \renewcommand\footnoterule{\kern -1ex}
 \renewcommand{\arraystretch}{1.3}
 \begin{tabular}{l*{5}{S}}
 \hline 
 \multicolumn{4}{c}{ADD (Higher is better)} \\
 \hline
 \multicolumn{1}{c}{Object} & 
 \multicolumn{1}{c}{T6D-Direct} & 
 \multicolumn{1}{c}{PoseCNN}  & 
 \multicolumn{1}{c}{TransPose}
\\\midrule

mug   & 72.1 & 57.7  & \multicolumn{1}{c}{75.7} \\
tuna fish can   & 59.0 & 70.4 & \multicolumn{1}{c}{60.2}\\
sugar box   & 81.8 & 68.6  & \multicolumn{1}{c}{84.5}\\
\textcolor{green}{bowl}   & 91.6 & 69.7  & \multicolumn{1}{c}{89.7} \\
master chef can  & 61.5 & 50.9  & \multicolumn{1}{c}{63.4}\\
tomato soup can   & 72.0 &  66.0  & \multicolumn{1}{c}{75.6} \\
\textcolor{green}{wood block}   & 90.7 & 65.8 & \multicolumn{1}{c}{90.7} \\
pudding box  & 72.7 & 62.9  & \multicolumn{1}{c}{78.3} \\
banana  & 87.4 & 91.3  & \multicolumn{1}{c}{90.4}\\
bleach cleanser & 65.0 & 50.5  & \multicolumn{1}{c}{70.2}\\
\hline
\multicolumn{4}{c}{ADD-S (Higher is better)} \\
 \hline
 mug & 89.8 & 78.0 &  90.1\\
tuna fish can  & 92.2 & 87.9 &  91.7\\
sugar box    & 90.3 & 84.3 &  93.1\\
\textcolor{green}{bowl}    & 91.6 & 69.7 &  92.3\\
master chef can   & 91.9 & 84.0 &  92.4\\
tomato soup can   & 88.9 & 80.9 &  90.8\\
\textcolor{green}{wood block}   & 90.7 & 65.8 &  90.6\\
pudding box   & 85.1 & 79.0 &  88.1\\
banana  & 93.8 & 85.9 &  94.5\\
bleach cleanser & 83.0 & 71.9 &  84.3\\
\hline
\textbf{Mean} (ADD) & 75.38 & 65.38 & 77.87\\
\hline
\textbf{Mean} (ADD-S) & 88.50 & 80.44 & 91.52\\

 \end{tabular}
 \end{table}
We can see that our proposed solution outperforms the other methods considering the ADD metric for all the objects except the "tuna fish can", "bowl", "wood block" and "banana" where our network closely compares with the other methods. Similarly, using the ADD-S metric, our solution outperforms the other methods except for the objects "tuna fish can" and "wood block". 

A similar comparison is conducted for our fruit dataset using the ADD and ADD-S metric as shown in Table \ref{tab3}. 
\begin{table}[h!]
 \sisetup{
 table-number-alignment = center,
 table-figures-integer = 1,
 table-figures-decimal = 4
 }
 \caption{Pose estimation Comparison using various approaches on Fruit Dataset}
 \label{tab3}
 \centering
 \renewcommand\footnoterule{\kern -1ex}
 \renewcommand{\arraystretch}{1.3}
 \begin{tabular}{l*{5}{S}}
 \hline 
 \multicolumn{4}{c}{ADD (Higher is better)} \\
 \hline
 \multicolumn{1}{c}{Object} & 
 \multicolumn{1}{c}{T6D-Direct} & 
 \multicolumn{1}{c}{PoseCNN}  & 
 \multicolumn{1}{c}{TransPose}
\\\midrule

Apple   & 78.7 & 62.4  & \multicolumn{1}{c}{82.4}\\
Avocado   & 81.3 & 71.4 & \multicolumn{1}{c}{82.6} \\
Banana   & 90.4 & 76.6  & \multicolumn{1}{c}{92.4} \\
Orange  & 71.4 & 59.7  & \multicolumn{1}{c}{79.3}\\ 
Lemon  & 89.5 & 71.9  & \multicolumn{1}{c}{89.8}\\
\hline
\multicolumn{4}{c}{ADD-S (Higher is better)} \\
 \hline
Apple   & 87.5 & 73.2 &  89.7\\
Avocado   & 86.2 & 82.9 &  92.6\\
Banana   & 92.7 & 82.3 &  93.2\\
Orange   & 84.6 & 80.2 &  87.8\\
Lemon  & 91.5 & 83.6 &  94.3\\
\hline
\textbf{Mean} (ADD) & 82.26 & 68.40 & 85.30\\
\hline
\textbf{Mean} (ADD-S) & 89.73 & 78.74 & 90.79\\

 \end{tabular}
 \end{table}
The mean from Table \ref{tab2} and Table \ref{tab3} shows the overall performance of TransPose across the sample objects. From the mean ADD and ADD-S, we can see that the depth refinement module improves the performance of 6D pose estimation. 

\section{Conclusion}
This paper proposes TransPose, an improved transformer-based 6D pose estimation network that utilises a depth refinement module to improve the overall solution performance. In contrast to other multi-modal networks that require more than one sensor and data type, TransPose utilises an RGB image for the 6D pose estimation and the depth refinement with the aid of a depth estimation network. The 6D poses are directly regressed by means of a proposed transformer network and further refined with a depth network. We compare our results of the depth network with other methods using the standard evaluation metrics. The performance of the depth network satisfies the purpose of 6D pose refinement. The results obtained using the standard evaluation metrics show a competitive depth outcome. We evaluate our results on multiple datasets for depth estimation and final object 6D pose regression. We extended the scope to a fruit dataset to prove the effectiveness of this pipeline in precision agriculture, particularly fruit picking. In the future, we aim at exploring the real-time onboard deployment of TransPose in conjunction with a robotics manipulator for real-time fruit picking application.

\bibliographystyle{IEEEtran}
\bibliography{references}
\end{document}